\documentclass{article}

\pdfoutput=1 

\usepackage{amsmath}
\usepackage{amssymb}
\usepackage{tikz}
\usepackage{rotating}
\usepackage{url}
\usepackage{xr}
\usepackage{geometry}
\usepackage{pdflscape}
\usepackage{xcolor,colortbl}

\definecolor{Gray}{gray}{0.9}
\newcolumntype{a}{>{\columncolor{Gray}}c}
\newcolumntype{b}{>{\columncolor{white}}c}

\usepackage{multirow}

\DeclareMathOperator*{\argmin}{arg\,min}
\DeclareMathOperator*{\argmax}{arg\,max}

\newcommand{\ndatasets}{$22$}
\newcommand{\nalgorithms}{$23$}
\newcommand{\timelimit}{$40$ minutes}
\newcommand{\memlimit}{$4$ GB}
\newcommand{\nseeds}{$10$}
\newcommand{\ncentset}{\{100,1000\}}

\newcommand{\norm}[1]{\left\lVert#1\right\rVert}
\usepackage{xcolor,colortbl}

\newcounter{nbdrafts}
\makeatletter
\newcommand{\checknbdrafts}{
\@latex@warning@no@line{****************************************************}
\ifnum \thenbdrafts > 0
\@latex@warning@no@line{* The document contains \thenbdrafts \space todo note(s)}
\else
\@latex@warning@no@line{* NO DRAFT ANYMORE}
\fi
\@latex@warning@no@line{****************************************************}}

\usepackage{times}
\usepackage{graphicx}
\usepackage{subfigure} 
\usepackage{natbib}

\usepackage[accepted]{icml2016} 
\icmltitlerunning{Fast K-Means with Accurate Bounds}

\begin{document} 

\twocolumn[
\icmltitle{Fast K-Means with Accurate Bounds}


\icmlauthor{James Newling}{james.newling@idiap.ch}
\icmladdress{Idiap Research Institute \& EPFL, 
            Switzerland}
\icmlauthor{Fran\c cois Fleuret}{francois.fleuret@idiap.ch}
\icmladdress{Idiap Research Institute \& EPFL,
            Switzerland}

\icmlkeywords{$k$-means, clustering, mini-batch, Elkan, YinYang, triangle inequality, exact, bounds, ICML}

\vskip 0.3in
]

\begin{abstract}
We propose a novel accelerated exact $k$-means algorithm, which outperforms the current state-of-the-art low-dimensional algorithm in 18 of 22 experiments, running up to 3$\times$ faster. We also propose a general improvement of existing state-of-the-art accelerated exact $k$-means algorithms through better estimates of the distance bounds used to reduce the number of distance calculations, obtaining speedups in 36 of 44 experiments, of up to 1.8$\times$. We have conducted experiments with our own implementations of existing methods to ensure homogeneous evaluation of performance, and we show that our implementations perform as well or better than existing available implementations. Finally, we propose simplified variants of standard approaches and show that they are faster than their fully-fledged counterparts in 59 of 62 experiments.
\end{abstract}


\section{Introduction}
The $k$-means problem is to compute a set of $k$ centroids to minimise the sum over data-points of
the squared distance to the nearest centroid. It is an NP-hard problem for which various effective approximation algorithms exist. The most popular is often referred to as Lloyd's algorithm, or simply as \textit{the} $k$-means algorithm. It has applications in data compression, data classification, density estimation and many other areas, and was recognised in~\citet{wu2008wu} as one of the top-10 algorithms in data mining. 

Lloyd's algorithm relies on a two-step iterative process: In the \textit{assignment} step, each sample is assigned to the cluster whose centroid is nearest. In the \textit{update} step, cluster centroids are updated in accordance with their assigned samples. Lloyd's algorithm is also called the \textit{exact} $k$-means algorithm, as there is no approximation in either of the two steps. This name can lead to confusion as the algorithm does not solve the $k$-means problem exactly.

The linear dependence on the number of clusters, the number of samples and the dimension of the space, means that Lloyd's algorithm requires upwards of a billion floating point operations per round on standard datasets such as those used in our experiments (\S\ref{sec:experiments}). 
This, coupled with slow convergence and the fact that several runs are often performed to find improved solutions, can make it slow. 

Lloyd's algorithm does not state how the assignment and update steps should be performed, and as such provides a scaffolding on which more elaborate algorithms can be constructed. These more elaborate algorithms, often called \textit{accelerated} exact $k$-means algorithms, are the primary focus of this paper. They can be dropped-in wherever Lloyd's algorithm is used.

\subsection{Approximate $k$-means}
Alternatives to exact $k$-means have been proposed. Certain of these rely on a relaxation of the assignment step, for example by only considering certain clusters according to some hierarchical ordering~\citep{Nister:2006:SRV:1153171.1153548}, or by using an approximate nearest neighbour search as in~\citet{Philbin07}. Others rely on a relaxation of the update step, for example by using only a subset of data to update centroids~\citep{Frahling:2006:FKI:1137856.1137879, Sculley:2010:WKC:1772690.1772862}.

When comparing approximate $k$-means clustering algorithms such as those just mentioned, the two criteria of interest are the quality of the final clustering, and the computational requirements. The two criteria are not independent, making comparison between algorithms difficult and often preventing their adoption. When comparing accelerated exact $k$-means algorithms on the other hand,  all algorithms produce the same final clustering, and so comparisons can be made based on speed alone. Once an accelerated exact $k$-means algorithm has been confirmed to provide a speed-up, it is rapidly adopted, automatically inheriting the trust which the exact algorithm has gained through its simplicity and extensive use over several decades.

\subsection{Accelerated Exact $k$-means}
The first published accelerated $k$-means algorithms borrowed techniques used to accelerate the nearest neighbour search. Examples are the adaptation of the algorithm of~\citet{Orchard} in~\citet{Phillips}, and the use of kd-trees~\citep{Bentley:1975:MBS:361002.361007} in~\citet{kanungo_2002_kmeans}. These algorithms relied on storing centroids in special data 
structures, enabling nearest neighbor queries to be processed without computing distances to all $k$ centroids.

The next big acceleration~\citep{elkan_2003_kmeansicml} came about by maintaing bounds on distances between samples and centroids, frequently resulting in more than 90\% of distance calculations being avoided. It was later shown~\citep{hamerly_2010_kmeans} that in low-dimensions, it is more effective to keep bounds on distances to only the two nearest centroids, and that in general bounding-based algorithms are significantly faster than tree-based ones. Further bounding-based algorithms were proposed by~\citet{drake_2013_masters} and ~\citet{ding_2015_yinyang}, each providing accelerations over their predecessors in certain settings. In this paper, we continue in the same vain.

\subsection{Our Contribution}
Our first contribution (\S\ref{sec:exponion}) is a new bounding-based accelerated exact $k$-means algorithm, the Exponion algorithm. Its closest relative is the Annular algorithm~\citep{drake_2013_masters}, which is the current state-of-the-art accelerated exact $k$-means algorithm in low-dimensions. We show that the Exponion algorithm is significantly faster than the Annular algorithm on a majority of low-dimensional datasets.

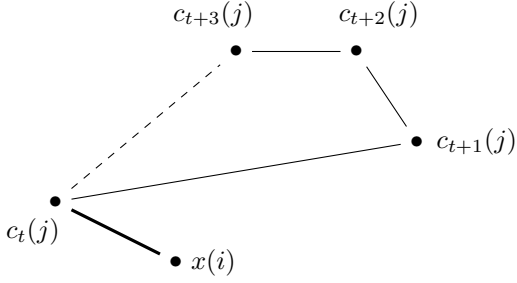
\begin{figure}
\center
\tikzstyle{position}=[]
\begin{tikzpicture}[scale=0.8]
\path node (x)  at ( 0.00, -1.00) [position] {$\bullet$}
      node (c0) at (-2.00,  0.00) [position] {$\bullet$}
      node (c1) at ( 4.00,  1.00) [position] {$\bullet$}
      node (c2) at ( 3.00,  2.50) [position] {$\bullet$}
      node (c3) at ( 1.00,  2.50) [position] {$\bullet$};

\node [xshift=5mm]              at (x) {$x(i)$};
\node [xshift=-3mm, yshift=-4mm] at (c0) {$c_t(j)$};
\node [xshift=8mm]               at (c1) {$c_{t+1}(j)$};
\node [xshift=3mm,yshift=5mm]    at (c2) {$c_{t+2}(j)$};
\node [xshift=-3mm,yshift=5mm]   at (c3) {$c_{t+3}(j)$};

\draw [thin] (c0) to (c1);
\draw [thin] (c1) to (c2);
\draw [thin] (c2) to (c3);
\draw [dashed] (c0) to (c3);
\draw [very thick] (x) to (c0);

\end{tikzpicture}
\caption{The classical sn-bound is the sum of the last known distance
  between the sample to a previous position of the centroid (thick
  solid line), with all the distances between successive positions of
  the centroid since then (thin solid lines). The ns-bound we propose uses the actual distance
  between that previous location of the centroid and its current one
  (dashed line).}  
\label{figure}
\end{figure}

Our second contribution (\S\ref{sec:improving-bounds}) is a technique for making bounds tighter, allowing further redundant distance calculations to be eliminated. The technique, illustrated in Figure~\ref{figure}, can be applied to all existing bounding-based $k$-means algorithms.

Finally, we show how certain of the current state-of-the-art algorithms can be accelerated through strict simplifications~(\S\ref{sec:simpl-elkans-algor} and \S\ref{sec:yinyang}). Fully parallelised implementations of all algorithms are provided under an open-source license at \url{https://github.com/idiap/eakmeans}

\section{Notation and baselines}
We describe four accelerated exact $k$-means algorithms in order of publication date. For two of these we propose simplified versions which offer natural stepping stones in understanding the full versions, as well being faster (\S\ref{res:simplifications}).

Our notation is based on that of~\citet{hamerly_2010_kmeans}, and only where necessary is new notation introduced. We use for example $N$ for the number of samples and $k$ for the number of clusters. Indices $i$ and $j$ always refer to data and cluster indices respectively, with a sample denoted by $x(i)$ and the index of the cluster to which it is assigned by $a(i)$. A cluster's centroid is denoted as $c(j)$.  We introduce new notation by letting $n_{1}(i)$ and $n_{2}(i)$ denote the indices of the clusters whose centroids are the nearest and second nearest to sample $i$ respectively.

Note that $a(i)$ and $n_1(i)$ are different, with the objective in a round of $k$-means being to set $a(i)$ to $n_1(i)$. $a(i)$ is a variable maintained by algorithms, changing within loops whenever a better candidate for the nearest centroid is found. On the other hand, $n_1(i)$ is introduced purely to aid in proofs, and is external to any algorithmic details. It can be considered to be the hidden variable which algorithms need to reveal.

All of the algorithms which we consider are elaborations of Lloyd's algorithm, and thus consist of repeating the assignment step and update step, given respectively as
\begin{align}
\label{eqn::dataupdate}
a(i) &\gets n_1(i), \ i \in \{1,\ldots, N\} \\
c(j) &\gets \frac{\sum_{i: a(i)=j} x(i)}{\|i: a(i) = j \|}, \ j \in \{1, \ldots, k\}. \label {eqn::cstep}
\end{align}
These two steps are repeated until there is no change to any $a(i)$, or some other stopping criterion is met. We reiterate that all the algorithms discussed provide the same output at each iteration of the two steps, differing only in how $a(i)$ is computed in~\eqref{eqn::dataupdate}.

\subsection{Standard algorithm (\texttt{sta})}
The Standard algorithm, henceforth \texttt{sta}, is the simplest implementation of Lloyd's algorithm. The only variables kept are $x(i)$ and $a(i)$ for $i \in \{1, \ldots, N\}$ and $c(j)$ for $j \in \{ 1,\ldots, k\}$. The assignment step consists of, for each $i$, calculating the distance from $x(i)$ to all centroids, thus revealing $n_1(i)$.

\subsection{Simplified Elkan's algorithm (\texttt{selk})}\label{sec:simpl-elkans-algor}
Simplified Elkan's algorithm, henceforth \texttt{selk}, uses a strict subset of the strategies described in~\citet{elkan_2003_kmeansicml}. In addition to $x(i)$, $a(i)$ and $c(j)$, the variables kept are $p(j)$, the distance moved by $c(j)$ in the last \mbox{update step}, and bounds $l(i,j)$ and $u(i)$, maintained to satisfy,
\begin{align*}
&l(i,j) \le \|x(i) - c(j)\|, & u(i) \ge \|x(i) - c(a(i))\|.
\end{align*}
These bounds are used to eliminate unnecessary centroid-data distance calculations using,
\begin{align}
u(i) < l(i,j) &\implies \|x(i) - c(a(i))\| < \|x(i) - c(j)\| \notag   \\
&\implies j \not= n_1(i).
\label{eqn::basicelkantrick}
\end{align}
We refer to~\eqref{eqn::basicelkantrick} as an \textit{inner} test, as it is performed within a loop over centroids for each sample. This as opposed to an \textit{outer} test which is performed just once per sample, examples of which will be presented later.

To maintain the correctness of the bounds when centroids move, bounds are updated at the beginning of each assignment step with
\begin{align}
\label{eqn::elkup}
&l(i,j) \gets l(i,j) - p(j), & u(i) \gets u(i) + p(a(i)).
\end{align}
The validity of these updates is a simple consequence of the triangle inequality (proof in~\ref{sec::supp::selk1}). We say that a bound is \textit{tight} if it is known to be equal to the distance it is bounding, a \textit{loose} bound is one which is not tight. For \texttt{selk}, bounds are initialised to be tight, and tightening a bound evidently costs one distance calculation.

When in a given round $u(i) \ge l(i,j)$, the test~\eqref{eqn::basicelkantrick} fails. The first time this happens, both $u(i)$ and $l(i,j)$ are loose due to preceding bound updates of the form~\eqref{eqn::elkup}. Tightening either bound may result in the test succeeding. Bound $u(i)$ should be tightened before $l(i,j)$, as it reappears in all tests for sample $i$ and will thus be reused. In the case of a test failure with tight $u(i)$ and loose $l(i,j)$ we tighten $l(i,j)$. A test failure with $u(i)$ and $l(i,j)$ both tight implies that centroid $j$ is nearer to sample $i$ than the currently assigned cluster centroid, and so $a(i)\gets j$ and $u(i)\gets l(i,j)$.

\subsection{Elkan's algorithm (\texttt{elk})}
\label{sec::elk}
The fully-fledged algorithm of~\citet{elkan_2003_kmeansicml}, henceforth \texttt{elk}, adds to \texttt{selk} an additional strategy for eliminating distance calculations in the assignment step. Two further variables, $cc(j,j')$, the matrix of inter-centroid distances, and $s(j)$, the distance from centroid $j$ to its nearest other centroid, are kept. A simple application of the triangle inequality, shown in~\ref{sec::supp::elkcc}, provides the following test,
\begin{equation}
\label{eqn::ccloctest}
\frac{cc(a(i), j)}{2} > u(i) \implies j \not= n_1(i).
\end{equation}
\texttt{elk} uses~\eqref{eqn::ccloctest} in unison with~\eqref{eqn::basicelkantrick} to obtain an improvement on the test of \texttt{elk}, of the form,
\begin{equation}
\label{eqn::elkinnerbound}
\max\left(l(i,j), \frac{cc(a(i), j)}{2}\right) > u(i) \implies j \not= n_1(i).
\end{equation}
In addition to the inner test~\eqref{eqn::elkinnerbound}, \texttt{elk} uses an outer test, whose validity follows from that of~\eqref{eqn::ccloctest}, given by,
\begin{equation}
\label{ccglobtest}
\frac{s(a(i))}{2} > u(i) \implies n_1(i) = a(i).
\end{equation}
If the outer test~\eqref{ccglobtest} is successful, one proceeds immediately to the next sample without changing $a(i)$.

\subsection{Hamerly's algorithm (\texttt{ham})}
The algorithm of~\citet{hamerly_2010_kmeans}, henceforth \texttt{ham}, represents a shift of focus from inner to outer tests, completely foregoing the inner test of \texttt{elk}, and providing an improved outer test.

The $k$ lower bounds per sample of \texttt{elk} are replaced by a single lower bound on all centroids other than the one assigned,
\begin{equation*}
l(i) \le \min_{j \not = a(i)}\|x(i) - c(j)\|.
\end{equation*}
The variables $p(j)$ and $u(i)$ used in \texttt{elk} have the same definition for \texttt{ham}. The test for a sample $i$ is
\begin{equation}
\label{alg::hamtest}
\max\left(l(i), \frac{s(a(i))}{2}\right) > u(i) \implies n_1(i) = a(i),
\end{equation}
with the proof of correctness being essentially the same as that for the inner test of \texttt{elk}. If test~\eqref{alg::hamtest} fails, then $u(i)$ is made tight. If test~\eqref{alg::hamtest} fails with $u(i)$ tight, then the distances from sample $i$ to all centroids are computed, revealing $n_1(i)$ and $n_2(i)$, allowing the updates $a(i) \gets n_1(i)$, $u(i) \gets \|x(i) - c(n_1(i))\|$ and $l(i) \gets \|x(i) - c(n_2(i))\|$. As with \texttt{elk}, at the start of the assignment step, bounds need to be adjusted to ensure their correctness following the update step. This is done via,
\begin{align*}
&l(i) \gets l(i) -  \argmax_{j\not=a(i)}{p(a(i))},& u(i) \gets u(i) + p(a(i)).
\end{align*}

\subsection{Annular algorithm (\texttt{ann})}
The Annular algorithm of~\citet{drake_2013_masters}, henceforth \texttt{ann}, is a strict extension of \texttt{ham}, adding one novel test. In addition to the variables used in \texttt{ham}, one new variable $b(i)$ is required, which roughly speaking is to $n_2(i)$ what $a(i)$ is to $n_1(i)$. Also, the centroid norms $\|c(j)\|$ should be computed and sorted in each round.

Upon failure of test~\eqref{alg::hamtest} with tight bounds in \texttt{ham}, $\|x(i) - c(j)\|$ is computed for all $j \in \{1, \ldots,k\}$ to reveal $n_1(i)$ and $n_2(i)$. With \texttt{ann}, certain of these $k$ calculations can be eliminated. Define the radius, and corresponding set of cluster indices,
\begin{align}
R(i) &= \max\left(u(i), \|x(i) - c(b(i))\|\right), \notag \\
\mathcal{J}(i) &= \{j: |\|c(j)\| - \|x(i)\|| \le R(i) \}. \label{eqn::Jinann}
\end{align}
The following implication (proved in~\ref{sec::supp::annulus}) is used 
\begin{equation*}
j \not\in \mathcal{J}(i) \implies j \not\in \{ n_1(i), n_2(i) \}.
\end{equation*}
Thus only distances from sample $i$ to centroids of the clusters whose indices are in $\mathcal{J}(i)$ need to be calculated for $n_1(i)$ and $n_2(i)$ to be revealed. Once $n_1(i)$ and $n_2(i)$ revealed, $a(i)$, $u(i)$ and $l(i)$ are updated as per \texttt{ham}, and \mbox{$b(i) \gets n_2(i)$}.

Note that by keeping an ordering of $\|c(j)\|$ the set $\mathcal{J}(i)$ can be determined in $\Theta(\log(K))$ operations with two binary searches, one for each of the inner and outer radii of $\mathcal{J}(i)$.

\subsection{Simplified Yinyang (\texttt{syin}) and Yinyang (\texttt{yin}) algorithms}\label{sec:yinyang}
The basic idea with the Yinyang algorithm~\citep{ding_2015_yinyang} and the Simplified Yinyang algorithm, henceforth \texttt{yin} and \texttt{syin} respectively, is to maintain consistent lower bounds for groups of clusters as a compromise between the $k-1$ lower bounds of \texttt{elk} and the single lower bound of \texttt{ham}. In~\citet{ding_2015_yinyang} the number of groups is fixed at one tenth the number of centroids. The groupings are determined and fixed by an initial clustering of the centroids. The algorithm appearing in the literature most similar to \texttt{yin} is Drake's algorithm of~\citep{drake_2012_original}, not to be confused with \texttt{ann}. According to~\citet{ding_2015_yinyang}, Drake's algorithm does not perform as well as \texttt{yin}, and we thus choose not to consider it in this paper.  

Denote by $G$ the number of groups of clusters. Variables required in addition to those used in \texttt{sta} are $p(j)$ and $u(i)$, as per \texttt{elk}, $\mathcal{G}(f)$, the set of indices of clusters belonging to the $f$'th group, $g(i)$, the group to which cluster $a(i)$ belongs, $q(f) = \max_{j \in \mathcal{G}}p(j)$, and bound $l(i, f)$, maintained to satisfy,
\begin{equation*}
l(i,f) \le \argmin_{j \in \mathcal{G}(f) \setminus \{ a(i) \}}\|x(i) - c(j)\|.
\end{equation*}
For both \texttt{syin} and \texttt{yin}, both an outer test and group tests are used. To these, \texttt{yin} adds an inner test. The outer test is
\begin{equation}
\label{eqn::obtgroup}
\min_{f \in  \{1, \ldots, G\}} l(i,f) > u(i) \implies a(i) = n_1(i).
\end{equation}
If and when test~\eqref{eqn::obtgroup} fails, group tests of the form
\begin{equation}
\label{eqn::ibtgroup}
l(i,f) > u(i) \implies a(i) \not\in\mathcal{G}(f),
\end{equation}
are performed. As with \texttt{elk} and \texttt{ham}, if test~\eqref{eqn::ibtgroup} fails with $u(i)$ loose, $u(i)$ is made tight and the test reperformed.

The difference between \texttt{syin} and \texttt{yin} arises when~\eqref{eqn::ibtgroup} fails with $u(i)$ tight. With \texttt{syin}, the simple approach of computing distances from $x(i)$ to all centroids in $\mathcal{G}(f)$, then updating $l(i,f), l(i, g(i)), u(i), a(i)$ and $g(i)$ as necessary, is taken. With \texttt{yin} a final effort at eliminating distance calculations by the use of a local test is made, as described in~\ref{sec::supp::yinyang}. As will be shown (\S\ref{res:simplifications}), it is not clear that the local test of \texttt{yin} makes it any faster. Finally, we mention how $u(i)$ and $l(i,f)$ are updated at the beginning of the assignment step for the \texttt{syin} and \texttt{yin},
\begin{align*}
l(i,f) &\gets l(i,f) - \argmax_{j\in{\mathcal{G}(f)}}{p(a(i))},  \\
u(i) &\gets  u(i) + p(a(i)).
\end{align*}

\section{Contributions and New Algorithms}
We first present (\S\ref{sec:exponion}) an algorithm which we call Exponion, and then (\S\ref{sec:improving-bounds}) an improved bounding approach.
\subsection{Exponion algorithm (\texttt{exp})}
\label{sec:exponion}

Like \texttt{ann}, \texttt{exp} is an extension of \texttt{ham} which adds a test to filter out $j \not\in \{n_1(i), n_2(i)\}$ when test~\eqref{alg::hamtest} fails. Unlike \texttt{ann}, where the filter is an origin-centered annulus, \texttt{exp} has as filter a ball centred on centroid $a(i)$. This change is motivated by the ratio of volumes of an annulus of width $r$ at radius $w$ and a ball of radius $r$ from the origin, which is $d \left(\frac{w}{r}\right)^{d-1}$ in $\mathbb{R}^d$. We expect $r$ to be greater than $w$, whence the expected improvement. 
Define,
\begin{align}
R(i) &= 2u(i) + s(a(i)), \notag \\
\mathcal{J}(i) &= \{j: \|c(j) - c(a(i))\| \le R(i) \}. \label{eqn::Jinexp}
\end{align}
The underlying test used (proof in~\ref{sec::supp::exponion}) is
\begin{equation*}
j \not\in \mathcal{J}(i) \implies j \not\in \{ n_1(i), n_2(i) \}.
\end{equation*}
In moving from \texttt{ann} to \texttt{exp}, the decentralisation from the origin to the centroids incurs two costs, one which can be explained algorithmically, the other is related to cache memory.

Recall that \texttt{ann} sorts $\|c(j)\|$ in each round, thus guaranteeing that the set of candidate centroids~\eqref{eqn::Jinann} can be obtained in $O(\log(k))$ operations. To guarantee that the set of candidate centroids~\eqref{eqn::Jinexp} can be obtained with $O(\log(k))$ operations requires that $\|c(j) - c(a(i))\|$ be sorted. For this to be true for all samples requires sorting $\|c(j) - c(j')\|$ for all $j \in \{1, \ldots, k\}$, increasing the overhead of sorting from $O(k \log k)$ to $O(k^2 \log k)$. 


The cache related cost is that, unless samples are ordered by $a(i)$, the bisection search performed to obtain $\mathcal{J}(i)$ is done with a different row of $c(j,j')$ for each sample, resulting in cache memory misses.

To offset these costs, we replace the exact sorting of $cc$ with a partial sorting, paying for this approximation with additional distance calculations. 
We maintain, for each centroid, $\lceil \log_2 k \rceil$ concentric annuli, each succesive annulus containing twice as many centroids as the one interior to it. For cluster $j$, annulus $f \in \{1,\ldots,\lceil \log_2 k \rceil\} $ is defined by inner and outer radii $e(j, f-1)$ and $e(j, f)$, and a list of indices $w(j,f)$ with $|w(j,f)| = 2^f$, where 
\begin{equation*}
w(j,f) = \{j': e(j, f-1) < \|c(j') - c(j)\| \le e(j, f)  \}.
\end{equation*}
Note that $w(j,f)$ is not an ordered set, but there is an ordering between sets, 
\begin{align*}
j' \in w(j,f), j'' \in w(j,f+1) \implies& \\
 \|c(j') - c(j)\| &< \|c(j'') - c(j)\|.
\end{align*}
Given a search radius $R(i)$, without a complete ordering of $c(j,j')$ we cannot obtain $\mathcal{J}(i)$ in $O(\log(k))$ operations, but we can obtain a slightly larger set $\mathcal{J^*}(i)$ defined by
\begin{align*}
f^*(i) &=  \min \{f :  e(a(i),f) \ge R(i)\},\\  
\mathcal{J^*}(i) &= \bigcup_{f \le f^{*}(i)} w(j,f), 
\end{align*}

in $\log\log(k)$ operations. It is easy to see that $|\mathcal{J^*}(i)| \le 2|\mathcal{J}(i)| $, and so using the partial sorting cannot cost more than twice the number of distance calculations. 

\subsection{Improving bounds (sn to ns)}
\label{sec:improving-bounds}
In all the algorithms presented so far, upper bounds (lower bounds) are updated in each round with increments (decrements) of norms of displacements. If tests are repeatedly successful, these increments (decrements) accumulate. Consider for example the upper bound update,
\begin{equation*}
u_{t_0 + 1}(i) \gets u_{t_0}(i) +  p_{t_0}(a(i)),
\end{equation*}
where subscripts denote rounds. The upper bound after $\delta t$ such updates without bound tightening is
\begin{equation}
\label{eqn::ns}
u_{t_0 + \delta t}(i) = u_{t_0}(i) + \sum_{t' = t_0}^{t + \delta t -1}  p_{t'}(a(i)).
\end{equation}
The summation term is a (s)um of (n)orms of displacement, thus we refer to it as an sn-bound and to an algorithm using only such an update scheme as an sn-algorithm. An alternative upper bound at round $t_0+\delta t$ is,
\begin{align}
u_{t_0 + \delta t}(i) & = u_{t_0}(i) + \norm{\sum_{t' = t_0}^{t_0 + \delta t -1}  c_{t' + 1}(i) - c_{t'}(i)}, \notag \\
                         & = u_{t_0}(i) + \|c_{t_0+\delta t}(i) - c_{t_0}(i)\|. \label{eqn::sn}
\end{align}
Bound~\eqref{eqn::sn} derives from the (n)orm of a (s)um, and hence we refer to it as an ns-bound. An ns-bound is guaranteed to be tighter than its equivalent sn-bound by a trivial application of the triangle inequality (proved in ~\ref{sec::supp::nsusn}). We have presented an upper ns-bound, but lower ns-bound formulations are similar. In fact, for cases where lower bounds apply to several distances simultaneously, due to the additional operation of taking a group maximum, there are three possible ways to compute a lower bound, details in \S\ref{supp:sec:3pos}.

\subsection{Simplified Elkan's algorithm-ns (\texttt{selk-ns})}
\label{sec::seance}

In transforming an sn-algorithm into an ns-algorithm, additional variables need to be maintained. These include a record of previous centroids $C$, where $C(j, t) = c_t(j)$, and displacement of $c(j)$ with respect to previous centroids, $P(j, t) = \|c(j) - c_t(j)\|$. We no longer keep rolling bounds for each sample, instead we keep a record of when most recently bounds were made tight and the distances then calculated. For Simplified Elkan's Algorithm-ns, henceforth \texttt{selk-ns}, we define $T(i, j)$ to be the last time $\|x(i) - c(j)\|$ was calculated, with corresponding distance $l(i,j) = \|x(i) - c_{T(i,j)}(j)\|$. We emphasise that $l(i,j)$ is defined differently here to in \texttt{selk}, with $u(i)$ similarly redefined as $u(i) = \|x(i) - c_{T(i,a(i))}(a(i))\|$.

The underlying test is
\begin{align*}
u(i) + P(a(i), T(i, a(i))) < l(i,j) - &P(j, T(i,j)) \\
&\implies j \not= n_1(i). \notag
\end{align*}
As with \texttt{selk}, the first bound failure for sample $i$ results in $u(i)$ being updated, with subsequent failures resulting in $l(i,j)$ being updated to the current distance. In addition, when $u(i)$ ($l(i,j)$) is updated, $T(i, a(i))$ ($T(i, j)$) is set to the current round.

Due to the additional variables $C, P$ and $T$, the memory requirement imposed is larger with \texttt{selk-ns} than with \texttt{selk-sn}. Ignoring constants, in round $t$ the memory requirement assuming samples of size $O(d)$ is, 
\begin{equation*}
\texttt{mem}_{ns} = O(Nd + Nk + ktd),
\end{equation*}
where $x, l$ and $C$ are the principal contributors to the above three respective terms. \texttt{selk} consists of only the first two terms, and so when $t > N/\min(k,d)$,
the dominant memory consumer in \texttt{selk-ns} is the new variable $C$. To guarantee that $C$ does not dominate memory consumption, an sn-like reset is performed in rounds $\{t : t \equiv 0\mod( N/\min(k,d))\}$, consisting of the following updates,
\begin{align*}
u(i) &\gets u(i) + P(a(i), T(i, a(i))),\\
l(i,j) &\gets l(i,j) - P(j, T(i,j)),\\
T(i,j) &\gets t,
\end{align*}
and finally the clearing of $C$.

\subsection{Changing Bounds for Other Algorithms}

All sn- to ns- coversions are much the same as that described in Section~\ref{sec::seance}. We have implemented versions of \texttt{elk}, \texttt{syin} and \texttt{exp} using ns-bounds, which we refer to as  \texttt{elk-ns}, \texttt{syin-ns} and  \texttt{exp-ns} respectively. 

\section{Experiments and Results}
\label{sec:experiments}
Our first set of experiments are conducted using a single core. We first establish that our implementations of baseline algorithms are as fast or faster than existing implementations. Having done this, we consider the effects of the novel algorithmic contributions presented, simplification, the Exponion algorithm, and ns-bounding. The final set of experiments are conducted on multiple cores, and illustrate how all algorithms presented parallelise gracefully.

We compare \nalgorithms{} $k$-means implementations, including our own implementations of all algorithms described, original implementations accompanying the papers~\citep{hamerly_2010_kmeans, drake_2013_masters, ding_2015_yinyang}, and implementations in two popular machine learning libraries, VLFeat and mlpack. We use the following notation to refer to implementations: \{\texttt{codesource}-\texttt{algorithm}\}, where \texttt{codesource} is one of \texttt{bay}~\citep{baylorgit}, \texttt{mlp}~\citep{mlpack2013}, \texttt{pow}~\citep{Low+al:uai10graphlab}, \texttt{vlf}~\citep{vedaldi08vlfeat} and \texttt{own} (our own code), and \texttt{algorithm} is one of the algorithms described.

Unless otherwise stated, times are wall times excluding data loading. We impose a time limit of 40 minutes and a memory limit of 4 GB on all \{dataset, implementation, $k$, seed\} runs. If a run fails to complete in 40 minutes, the corresponding table entry is `t'. Similarly, failure to execute with 4GB of memory results in a table entry `m'. We confirm that for all \{dataset, $k$, seed\} triplets, all implementations which complete within the time and memory constraint take the same number of iterations to converge to a common local minimum, as expected. 

The implementations are compared over the \ndatasets{} datasets presented in Table~\ref{tab::datasets}, for $k \in \ncentset{} $, with \nseeds{} distinct centroid initialisations (seeds). For all $\{$dataset, $k$, seed$\}$ triplets, the \nalgorithms{} implementations are run serially on a machine with an Intel i7 processor and 8MB of cache memory. All experiments are performed using double precision floating point numbers.

\setlength{\tabcolsep}{1.2pt}
\begin{table}[ht!]
\centering
\begin{tabular}{ |c|c|c| } 
\hline
 & $d$ & $N$ \\
\hline
i & 2 & 100k \\
ii & 2 & 169k \\
iii & 2 & 1m \\
iv  & 3 & 165k \\
v  & 3 & 164k \\
\hline
\end{tabular}
\begin{tabular}{ |c|c|c| } 
\hline
 & $d$ & $N$ \\
\hline
vi & 4 & 200k \\
vii & 4 & 200k \\
viii & 9 & 68k \\
ix  & 11 & 41k \\
x  & 15 & 166k \\
xi & 17 & 23k \\
\hline
\end{tabular}
\begin{tabular}{ |c|c|c| } 
\hline
 & $d$ & $N$ \\
\hline
xii  & 28 & 66k \\
xiii & 30 & 1m \\
xiv  & 50 & 60k \\
xv  & 50 & 130k \\
xvi  & 55 & 581k \\
\hline
\end{tabular}
\begin{tabular}{ |c|c|c| } 
\hline
 & $d$ & $N$ \\
\hline
xvii  & 68 & 2.6m \\
xviii & 74 & 146k \\
xix  & 108 & 1m \\
xx  & 128 & 14k \\
xxi & 310 & 95k \\
xxii & 784 & 60k \\
\hline
\end{tabular}
\caption{The \ndatasets{} datasets used in experiments, ranging in dimension from 2 to 784. The datasets come from: the UCI, KDD and KEEL repositories (11,2,2), MNIST and STL-10 image databases (2,1), random (2), European Bioinformatics Institute (1) and Joensuu University (1). Full names and further details in~\ref{sec::supp::results}.}
\label{tab::datasets}
\end{table}

Findings in \citet{drake_2013_masters} suggest that the best algorithm to use for a dataset depends primarily on dimension, where in low-dimensions, \texttt{ham} and \texttt{ann} are fastest, in high-dimensions \texttt{elk} is fastest, and in intermediate dimensions an approach maintaining a fractional number of bounds, Drake's algorithm, is fastest. Our findings corroborate these on real datasets, although the lines separating the three groups are blurry.  In presenting our results we prefer to consider a partitioning of the datasets into just two groups about the dimension $d=20$. \texttt{ham} and its derivatives are considered for $d<20$, \texttt{elk} and its derivatives for $d \ge 20$, and \texttt{syin} and \texttt{yin} for all $d$.


\subsection{Single core experiments}

A complete presentation of wall times and number of iterations for all \{dataset, implementation, $k$\} triplets is presented over two pages in Tables~\ref{supp:tab:100} and~\ref{supp:tab:1000} (\S\ref{sec::supp::results}). Here we attempt to summarise our findings. We first compare implementations of published algorithms (\S\ref{res:baselines}), and then show how \texttt{selk} and \texttt{syin} often outperform their more complex counterparts (\S\ref{res:simplifications}). We show that \texttt{exp} is in general much faster than \texttt{ann} (\S\ref{sec:annvsexp}), and finally show how using ns-bounds can accelerate algorithms (\S\ref{res:snns}) .

\subsubsection{Comparing Implementations of Baselines}
\label{res:baselines}
There are algorithmic techniques which can speedup all $k$-means algorithms discussed in this paper, we mention a few which we use. One is pre-computing the squares of norms of all samples just once, and those of centroids once per round. Another, first suggested in~\citet{hamerly_2010_kmeans}, is to update the sum of samples by considering only those samples whose assignment changed in the previous round. A third optimisation technique is to decompose while-loops which contain inner branchings dependant on the tightness of upper bounds into separate while-loops, eliminating unnecessary comparisons. Finally, while there are no large matrix operations with bounding-based algorithms, in high-dimensions distance calculations can be accelerated by the use of SSE, as in VLFeat, or by fast implementations of BLAS, such as OpenBLAS~\cite{openblas}. 

Our careful attention to optimisation is reflected in Table~\ref{tab:basecompare} (\S\ref{app:tabs}), where implementations of \texttt{elk}, \texttt{ham}, \texttt{ann} and \texttt{yin} are compared. The values shown are ratios of mean runtimes using another implementation (column) and our own implementation of the same algorithm, on a given dataset (row). Our implementations are faster in all but 4 comparisons. 


\subsubsection{Benefits of Simplification}
\label{res:simplifications}
We compare published algorithms \texttt{elk} and \texttt{yin} with their simplified counterparts \texttt{selk} and \texttt{syin}.
The values in Table~\ref{tab:simplifying} are ratios of mean runtimes using simplified and original algorithms, values less than 1 mean that the simplified version is faster. We observe that \texttt{selk} is faster than \texttt{elk} in 16 of 18 experiments, and \texttt{syin} is faster than \texttt{yin} in 43 of 44 experiments, often dramatically so. 

It is interesting to ask why the inventors of \texttt{elk} and \texttt{yin} did not instead settle on algorithms \texttt{selk} and \texttt{syin} respectively. A partial answer might relate to the use of BLAS, as the speedup obtained by simplifying \texttt{yin} to \texttt{syin} never exceeds more than 10\% when BLAS is deactivated. \texttt{syin} is more responsive to BLAS than \texttt{yin} as it has larger matrix multiplications due to it not having a final filter.

\begin{table}[ht!]
\centering

\begin{tabular}{|c|cc|c|cc||c|cc|}
\hline
\multicolumn{6}{|c||}{ \texttt{own-yin} $\rightarrow$ } &
\multicolumn{3}{c|}{\texttt{own-elk} $\rightarrow$ } \\
\multicolumn{6}{|c||}{\texttt{own-syin} } &
\multicolumn{3}{c|}{\texttt{own-selk} } \\
\hline
 & 100 & 1000 &  & 100 & 1000 & & 100 & 1000 \\
\hline

i & 0.96 & 0.90 & xii & 0.58 & 0.76 & xii & 0.85 & \textit{1.05} \\ 
ii & \textit{1.03} & 0.86 & xiii & 0.66 & 0.61 & xiii & 0.97 & m \\ 
iii & 0.88 & 0.92 & xiv & 0.50 & 0.55 & xiv & 0.84 & 0.57 \\ 
iv & 0.94 & 0.87 & xv & 0.49 & 0.58 & xv & 0.54 & 0.49 \\ 
v & 0.93 & 0.88 & xvi & 0.49 & 0.66 & xvi & 0.92 & m \\ 
vi & 0.91 & 0.87 & xvii & 0.44 & 0.58 & xvii & 0.75 & m \\ 
vii & 0.96 & 0.90 & xviii & 0.42 & 0.47 & xviii & 0.86 & 0.66 \\ 
viii & 0.79 & 0.80 & xix & 0.36 & 0.42 & xix & 0.72 & m \\ 
ix & 0.77 & 0.80 & xx & 0.38 & 0.60 & xx & \textit{1.12} & 0.74 \\ 
x & 0.72 & 0.73 & xxi & 0.32 & 0.36 & xxi & 0.89 & 0.73 \\ 
xi & 0.64 & 0.71 & xxii & 0.36 & 0.38 & xxii & 0.99 & 0.89 \\ 

\hline
\end{tabular}
\caption{Comparing \texttt{yin} and \texttt{elk} to simplified versions \texttt{syin} and \texttt{selk}. Values are ratios of mean runtimes of simplified versions to their originals, for different low-dimensional datasets (rows) and $k$ (columns). Values less than 1 mean that the simplified version is faster. In all but 3 of 62 cases (italicised), simplification results in speedup, by as much as 3$\times$.}
\label{tab:simplifying}
\end{table}

\subsubsection{From Annular to Exponion}
\label{sec:annvsexp}
We compare the Annular algorithm (\texttt{ann}) with the Exponion algorithm (\texttt{exp}). The values in Table~\ref{tab:ann_exp} are ratios of mean runtimes (columns $q_t$) and of mean number of distance calculations (columns $q_{au}$). Values less than 1 denote better performance with \texttt{exp}. We observe that \texttt{exp} is markedly faster than \texttt{ann} on most low-dimensional datasets, reducing by more than 30\% the mean runtime in 17 of 22 experiments. The primary reason for the speedup is the reduced number of distance calculations. 


Table~\ref{tab:snwins} summarises how many times each of the sn-algorithms is fastest on the 44 \{dataset, $k$\} experiments, ns-algorithms excluded. The 13 experiments on which \texttt{exp} is fastest are all very low-dimensional ($d < 5$), the 24 on which \texttt{syin} is fastest are intermediate ($8 < d < 69$) and \texttt{selk} or \texttt{elk} are fastest in very high dimensions ($d > 73$). For a detailed comparison across all algorithms, consult Tables~\ref{supp:tab:100} and~\ref{supp:tab:1000} (\S\ref{sec::supp::results}).

\setlength{\tabcolsep}{2.7pt}
\begin{table}[ht!]
\centering

\begin{tabular}{|c|cc|cc|c|cc|cc|}

\hline
\multicolumn{10}{|c|}{ \texttt{own-ann} $\rightarrow$ \texttt{own-exp} } \\
\hline
& 
\multicolumn{2}{c|}{100} &
\multicolumn{2}{c|}{1000} & &
\multicolumn{2}{c|}{100} &
\multicolumn{2}{c|}{1000} \\
\hline
& $q_t$ & $q_{au}$ & $q_t$ & $q_{au}$ & 
& $q_t$ & $q_{au}$ & $q_t$ & $q_{au}$ \\
\hline

 i & 0.48 & 0.52 & 0.72 & 0.61  &  vii & 0.71 & 0.80 & 0.36 & 0.32 \\
 ii & 0.54 & 0.80 & 0.58 & 0.50  &  viii & \textit{1.12} & \textit{1.24} & \textit{1.02} & \textit{0.93} \\
 iii & 0.53 & 0.58 & 0.48 & 0.44  &  ix & 0.96 & 0.99 & 0.73 & 0.64 \\
 iv & 0.63 & 0.80 & 0.36 & 0.33  &  x & 0.67 & 0.65 & 0.55 & 0.41 \\
 v & 0.63 & 0.80 & 0.37 & 0.34  &  xi & \textit{1.24} & \textit{1.43} & \textit{1.30} & \textit{1.16} \\
 vi & 0.62 & 0.73 & 0.42 & 0.38  &   &  &  &  &  \\

\hline
\end{tabular}

\caption{ Ratios of mean runtimes (`$q_{t}$') and mean number of distance calculations (`$q_{au}$') using the Exponion (\texttt{own-exp}) and Annular (\texttt{own-ann}) algorithms, on datasets with $d < 20$. Exponion is faster in all but the four italicised cases. The speedup is primarily due to the reduced number of distance calculations.}
\label{tab:ann_exp}
\end{table}
\begin{table}[ht!]
\centering
\begin{tabular}{|ccccccc|}
\hline
\texttt{ham} & \texttt{ann} & \texttt{exp} & \texttt{syin} & \texttt{yin} & \texttt{selk}  & \texttt{elk} \\ 
\hline
0 & 0 & 13 & 24 & 0 & 6 & 1 \\
\hline
\end{tabular}
\caption{Number of times each sn-algorithm is fastest, over the 44 \{dataset, $k$\} experiments, ns-algorithms not considered here.}
\label{tab:snwins}
\end{table}


\subsubsection{From sn to ns Bounding}
\label{res:snns}
For each of the 44 \{dataset, $k$\} experiments, we compare the fastest sn-algorithm with its ns-variant. The results are presented in Table~\ref{tab:nssn}. Columns `\texttt{x}' denote the fastest sn-algorithm. Values are ratios of means over runs of some quantity using the ns- and sn- variants. The ratios are $q_{t}$ (runtimes), $q_{a}$ (number of distance calculations in the assignment step) and  $q_{au}$ (total number of distance calculations). 

In all but 8 of 44 experiments (italicised), we observe a speedup using ns-bounding, by up to 45\%. As expected, the number of distance calculations in the assignment step is never greater when using ns-bounds, however the total number of distance calculations is occasionally increased due to initial variables being maintained. 

\begin{table}[h!]
\centering

\begin{tabular}{|c|cccc|cccc|}
\hline
\multicolumn{9}{|c|}{ \texttt{own-x} $\rightarrow$ \texttt{own-x-ns} } \\ 
\hline
& \multicolumn{4}{c|}{ 100 } 
& \multicolumn{4}{c|}{ 1000 } \\
\hline
& \texttt{x} & $q_t$ & $q_a$ & $q_{au}$ 
& \texttt{x} & $q_t$ & $q_a$ & $q_{au}$  \\
\hline

i  & \texttt{exp}  &  0.96  &  0.97  &  0.99 & \texttt{exp}  &  0.99  &  0.98  &  1.00\\ 
ii  & \texttt{exp}  &  0.94  &  0.97  &  0.97 & \texttt{exp}  &  0.99  &  0.99  &  1.00\\ 
iii  & \texttt{exp}  &  0.95  &  0.97  &  0.98 & \texttt{exp}  &  0.98  &  0.96  &  0.99\\ 
iv  & \texttt{exp}  &  0.97  &  0.97  &  0.97 & \texttt{exp}  &  0.99  &  0.97  &  0.98\\ 
v  & \texttt{exp}  &  0.96  &  0.97  &  0.97 & \texttt{exp}  &  0.98  &  0.96  &  0.98\\ 
vi  & \texttt{exp}  &  0.95  &  0.96  &  0.97 & \texttt{exp}  &  0.97  &  0.96  &  0.98\\ 
vii  & \texttt{syin}  &  0.98  &  0.82  &  0.86 & \texttt{exp}  &  0.98  &  0.96  &  0.99\\ 
viii  & \texttt{syin}  &  0.98  &  0.86  &  0.88 & \texttt{syin}  &  0.87  &  0.44  &  0.65\\ 
ix  & \texttt{syin}  &  0.98  &  0.83  &  0.86 & \texttt{syin}  &  0.83  &  0.32  &  0.66\\ 
x  & \texttt{syin}  &  \textit{1.03}  &  \textit{0.91}  &  \textit{0.92} & \texttt{syin}  &  \textit{1.11}  &  \textit{0.72}  &  \textit{0.80}\\ 
xi  & \texttt{selk}  &  0.92  &  0.80  &  0.84 & \texttt{syin}  &  0.81  &  0.56  &  0.69\\ 
xii  & \texttt{syin}  &  \textit{1.00}  &  \textit{0.86}  &  \textit{0.88} & \texttt{syin}  &  0.96  &  0.51  &  0.85\\ 
xiii  & \texttt{syin}  &  0.96  &  0.84  &  0.84 & \texttt{syin}  &  0.87  &  0.58  &  0.61\\ 
xiv  & \texttt{syin}  &  0.99  &  0.86  &  0.87 & \texttt{syin}  &  0.74  &  0.51  &  0.63\\ 
xv  & \texttt{syin}  &  \textit{1.06}  &  \textit{0.93}  &  \textit{0.94} & \texttt{syin}  &  0.94  &  0.58  &  0.69\\ 
xvi  & \texttt{syin}  &  \textit{1.04}  &  \textit{0.91}  &  \textit{0.93} & \texttt{syin}  &  0.98  &  0.61  &  0.79\\ 
xvii  & \texttt{syin}  &  1.00  &  0.87  &  0.89 & \texttt{syin}  &   m & m & m \\ 
xviii  & \texttt{selk}  &  0.89  &  0.81  &  0.82 & \texttt{syin}  &  0.75  &  0.64  &  0.68\\ 
xix  & \texttt{selk}  &  0.88  &  0.84  &  0.85 & \texttt{syin}  &  0.91  &  0.75  &  0.77\\ 
xx  & \texttt{elk}  &  \textit{1.02}  &  \textit{0.96}  &  \textit{1.02} & \texttt{selk}  &  \textit{1.06}  &  \textit{0.99}  &  \textit{1.00}\\ 
xxi  & \texttt{selk}  &  0.85  &  0.81  &  0.82 & \texttt{selk}  &  0.72  &  0.72  &  0.73\\ 
xxii  & \texttt{selk}  &  0.80  &  0.77  &  0.78 & \texttt{selk}  &  0.55  &  0.68  &  0.69\\

\hline
\end{tabular}

\caption{The effect of using ns-bounds. Columns `\texttt{x}' denotes the fastest sn-algorithm for a particular  \{dataset, $k$\} experiment. Columns `$q_{t}$' denote the ratio of mean runtimes of ns- and sn- variants of \texttt{x}. Italicised values are cases where using ns-bounding results in a slow down ($q_{t} > 1$), in the majority of cases there is a speedup. `$q_{a}$' and `$q_{au}$' denote ratios of ns- to sn- mean number of distance calculations in the assignment step (\textit{a})  and in total (\textit{au}). `m' described in paragraph 3 of \S\ref{sec:experiments}.}
\label{tab:nssn}
\end{table}

\subsection{Multicore experiments}
We have implemented parallelised versions of all algorithms described in this paper using the C++11 thread support library. To measure the speedup using multiple cores, we compare the runtime using four threads to that using one thread on a non-hyperthreading four core machine. The results are summarised in Table~\ref{tab:pllintab}, where near fourfold speedups are observed.

\setlength{\tabcolsep}{1.9pt}
\begin{table}[ht!]
\centering
\begin{tabular}{|ccc|}
\hline
\multicolumn{3}{|c|}{i-xi} \\
\hline
& 100 & 1000 \\
\hline
\texttt{own-exp-ns} & 0.29 & 0.31 \\
\texttt{own-syin-ns}  & 0.31 & 0.29 \\
\hline
\end{tabular}
\begin{tabular}{|ccc|}
\hline
\multicolumn{3}{|c|}{xii-xxii} \\
\hline
& 100 & 1000 \\
\hline
\texttt{own-selk-ns} & 0.33 & 0.30 \\
\texttt{own-elk-ns}  & 0.30 & 0.28 \\
\texttt{own-syin-ns}  & 0.27 & 0.27 \\
\hline
\end{tabular}
\caption{The median speedup using four cores. The median is over i-xi on the left and xii-xxii on the right.}
\label{tab:pllintab}
\end{table}

\section{Conclusion and future work}
The experimental results presented show that the ns-bounding scheme makes exact $k$-means algorithms faster, and that our Exponion algorithm is significantly faster than existing state-of-the-art algorithms in low-dimensions. Both can be seen as good default choices for $k$-means clustering on large data-sets.

The main practical weakness that remains is the necessary prior selection of which algorithm to use, depending on the dimensionality of the problem at hand. This should be addressed through an adaptive procedure able to select automatically the optimal algorithm through an efficient exploration/exploitation strategy. The second and more prospective direction of work will be to introduce a sharing of information between samples, instead of processing them independently.

\section*{Acknowledgements}
James Newling was funded by the Hasler Foundation under the grant 13018 MASH2.

\appendix
\section{Table Comparing Implementations}
\label{app:tabs}
\setlength{\tabcolsep}{1.9pt}
\begin{table}[h!]
\centering

\begin{tabular}{ |c|c|cc|c|c||c|c|ccc| } \hline

&  \begin{sideways} \texttt{ } \, \end{sideways}
&  \begin{sideways} \texttt{bay-ham} \, \end{sideways}
&  \begin{sideways} \texttt{mlp-ham} \, \end{sideways}
&  \begin{sideways} \texttt{bay-ann} \, \end{sideways}
&  \begin{sideways} \texttt{pow-yin} \, \end{sideways}
&  \begin{sideways} \texttt{ } \, \end{sideways}
&  \begin{sideways} \texttt{pow-yin} \, \end{sideways}
&  \begin{sideways} \texttt{bay-elk} \, \end{sideways}
&  \begin{sideways} \texttt{mlp-elk} \, \end{sideways}
&  \begin{sideways} \texttt{vlf-elk} \, \end{sideways}
\\
	
\hline		
\multirow{11}{*}{\begin{sideways} 100 \end{sideways}}
&   i & 1.23 & 1.04 & \textit{0.78} & 7.52 & xii & 2.51 & 2.69 & 1.87 & 2.48 \\
&   ii & 1.28 & \textit{0.99} & \textit{0.86} & 4.91 & xiii & 3.84 & 1.36 & 1.56 & t \\
&   iii & 1.19 & 1.27 & \textit{0.88} & 9.84 & xiv & 1.98 & 1.75 & 1.53 & 2.72 \\
&   iv & 1.59 & 1.15 & 1.24 & 6.21 & xv & 2.21 & 1.48 & 1.48 & 2.01 \\
&   v & 1.59 & 1.20 & 1.24 & 6.01 & xvi & 2.30 & 1.68 & 2.00 & 2.85 \\
&   vi & 1.78 & 1.27 & 1.32 & 6.41 & xvii & m & 1.79 & 1.88 & 2.61 \\
&   vii & 1.78 & 1.17 & 1.48 & 5.63 & xviii & 1.69 & 1.91 & 1.46 & 2.68 \\
&   viii & 2.67 & 1.38 & 2.38 & 3.99 & xix & 1.49 & 1.64 & 1.74 & 2.44 \\
&   ix & 2.93 & 1.51 & 2.90 & 3.65 & xx & 1.35 & 2.53 & 2.21 & 2.41 \\
&   x & 3.59 & 1.75 & 2.67 & 3.28 & xxi & 1.24 & 2.35 & 1.57 & 1.81 \\
&   xi & 3.89 & 2.04 & 3.18 & 2.17 & xxii & 1.16 & 2.86 & 1.43 & 1.35 \\
\hline		
\multirow{11}{*}{\begin{sideways} 1000 \end{sideways}}
&   i & 1.51 & 1.03 & 1.06 & 7.57 & xii & 3.37 & 6.21 & 3.20 & 2.44 \\
&   ii & 1.52 & 1.04 & 1.17 & 8.03 & xiii & t & m & m & m \\
&   iii & 1.47 & 1.05 & 1.04 & 8.57 & xiv & 2.09 & 1.89 & 1.86 & 2.07 \\
&   iv & 1.77 & 1.09 & 1.59 & 6.98 & xv & 3.14 & 1.43 & 2.76 & 1.80 \\
&   v & 1.77 & 1.09 & 1.59 & 7.01 & xvi & 3.98 & m & m & m \\
&   vi & 2.07 & 1.17 & 1.79 & 7.23 & xvii & m & m & m & m \\
&   vii & 1.99 & 1.17 & 1.73 & 6.57 & xviii & 1.82 & 1.78 & 1.40 & 1.92 \\
&   viii & 3.01 & 1.38 & 2.97 & 4.63 & xix & t & m & m & m \\
&   ix & 3.28 & 1.58 & 3.34 & 4.06 & xx & 2.06 & 6.17 & 2.60 & 1.72 \\
&   x & 3.92 & 1.76 & 3.57 & 5.08 & xxi & 1.32 & 2.88 & 1.80 & 1.51 \\
&   xi & 4.08 & 1.99 & 4.03 & 2.89 & xxii & 1.17 & 4.82 & 1.74 & 1.28 \\
\hline
\end{tabular}
\caption{Comparing implementations. For 100 (above) and 1000 (below) clusters, and in low- (left) and high- (right) dimensions. Existing implementations (colums) of \texttt{ham}, \texttt{ann}, \texttt{yin} and \texttt{elk} are compared to our implementations as a ratio of mean runtimes, with the mean runtime of our implementation in the denominator. Values greater than 1 mean our implementation runs faster. `t' and `m' are described in paragraph 3 of \S\ref{sec:experiments}.}
\label{tab:basecompare}
\end{table}


\checknbdrafts

\bibliography{kmeans}
\bibliographystyle{icml2016}

\onecolumn

\renewcommand{\thesection}{SM-\Alph{section}}

\section{Proofs}
We will use subscripts to denote rounds of $k$-means, and $B(x, r)$ to denote the closed ball centered on $x$ of radius $r$.

\subsection{Proof of correctness of Elkan's algorithm update}
\label{sec::supp::selk1}
By the definition of the lower bound update,
\begin{align*}
l_{t_0 + 1}(i,j) &= l_{t_0}(i,j) - p_{t_0}(j). 
\intertext{Using that $l_{t_0}$ is a valid bound, the definition of $p_{t_0}$, and the triangle inequality, }
&\le \|x(i) - c_{t_0}(j)\| - p_{t_0}(j), \\
&\le \|x(i) - c_{t_0}(j)\| - \|c_{t_0}(j) - c_{t_0 + 1}(j)\| \\
&\le \|x(i) - c_{t_0 + 1}(j)\|.
\end{align*}
Thus the lower bound update is valid. Similarly for the upper bound,
\begin{align*}
u_{t_0 + 1}(i,j) &= u_{t_0}(i) + p_{t_0}(a(i)),  \\
&\ge \|x(i) - c_{t_0}(a(i))\| + p_{t_0}(a(i)), \\
&\ge \|x(i) - c_{t_0}(a(i))\| + \\
&\;\;\;\;\;\;\; \|c_{t_0}(a(i)) - c_{t_0 + 1}(a(i))\|, \\
&\ge \|x(i) - c_{t_0 + 1}(a(i))\|.
\end{align*}
This proves that the upper bound update is valid.

\subsection{Proof of correctness of Elkan's algorithm intercentroid test}
\label{sec::supp::elkcc}
Suppose that,
\begin{equation*}  
\frac{cc(a(i), j)}{2} > u(i).
\end{equation*}
Then, by the triangle inequality and previous definitions,
\begin{align*}
\|c(j) - x(i)\| &\ge  \|c(j) - c(a(i))\| - \\
&\;\;\;\;\;\;\;\;\;\; \|c(a(i)) - x(i)\|, \\
&\ge cc(a(i), j) - u(i), \\
&\ge 2u(i) - u(i), \\
&\ge u(i). 
\end{align*}
Thus $c(a(i))$ is nearer to $x(i)$ than $c(j)$ is, and so  $j \not= n_1(i)$.

\subsection{Proof of correctness of Annular algorithm test}
\label{sec::supp::annulus}
Recall the definition of $R(i)$, 
\begin{equation*}
R(i) = \max\left(u(i), \|x(i) - c(b(i))\|\right). 
\end{equation*}
Following directly from this definition and the definition of $u(i)$, we have $c(a(i)), c(b(i)) \in B(x(i), R(i))$. Therefore by the definitions of $n_1(i)$ and $n_2(i)$, we have that $c(n_1(i)), c(n_2(i)) \in B(x(i), R(i))$. The triangle inequality now provides
\begin{equation}
\label{eqn::annfilter}
|\|c(j)\| - \|x(i)\|| > R(i) \implies \|c(j) - x(i)\| > R(i),
\end{equation}
Thus by the definition of $\mathcal{J}(i)$,
\begin{equation*}
\mathcal{J}(i) = \{j: |\|c(j)\| - \|x(i)\|| \le R(i) \},
\end{equation*}
we can say,
\begin{equation*}
j \not\in \mathcal{J}(i) \implies j \not\in \{ n_1(i), n_2(i) \}.
\end{equation*}

\subsection{Proof of correctness of Exponion algorithm test}
\label{sec::supp::exponion}




Let $nn(j) \in \{1,\ldots, k\} \setminus \{j\}$ denote the index the cluster whose centroid is nearest to the centroid of cluster $j$ other than $j$, that is the centroid at distance $s(j)$ from centroid $j$.

By definitions we have 
\begin{align*} 
c(a(i)) &\in B(x(i), u(i)),\\
c(nn(a(i))) &\in B(c(a(i)), s(a(i))).
\end{align*}
Combining these we have
\begin{equation} 
c(a(i)), c(nn(a(i))) \in  B(x(i), u(i) + s(a(i))), \label{eqn::twobirds}
\end{equation}
Basic geometric arguments provide
\begin{equation}
B\left(x(i), u(i) + s(a(i))\right) \subseteq B(c(a(i)), 2u(i) + s(a(i))).\label{eqn::supsersetting}
\end{equation}
From~\eqref{eqn::twobirds} we deduce that 
\begin{equation*}
c(n_1(i)), c(n_2(i)) \in  B(x(i), u(i) + s(a(i))),
\end{equation*}
and hence by~\eqref{eqn::supsersetting} we have 
\begin{equation*}
c(n_1(i)), c(n_2(i)) \in (c(a(i)), 2u(i) + s(a(i))),
\end{equation*} 
completing the proof.

\subsection{Proof that ns upper bound is tighter than sn upper bound}
\label{sec::supp::nsusn}
\begin{align*}
u^{ns}_{t_0 + \delta t}(i) &= u_{t_0}(i) + \norm{\sum_{t' = t_0}^{t_0 + \delta t -1}  c_{t' + 1}(i) - c_{t'}(i)},\\
&\le u_{t_0}(i) + \sum_{t' = t_0}^{t_0 + \delta t -1}  \| c_{t' + 1}(i) - c_{t'}(i) \|,\\
&\le u^{sn}_{t_0 + \delta t}(i).
\end{align*}

\section{Detailed descriptions}
\subsection{The inner Yinyang test}
\label{sec::supp::yinyang}
We need some temporary notation to present the test which the Yinyang algorithm employs,
\begin{align*}
j_1(i,f) &= \displaystyle \argmin_{j \in \mathcal{G}(f)} \|x(i) - c(j)\|, \\
j_2(i,f) &= \displaystyle  \argmin_{j \in \mathcal{G}(f) \setminus \{ j_1(f)\} } \|x(i) - c(j)\|, \\
r_2(i,f) &= \|x(i) - c(j_2(f))\|.
\end{align*}
The Yinyang test hinges on the fact that centroids in $\mathcal{G}(f)$ which lie beyond radius $r_2(i,f)$ of $x(i)$ do not affect the variable updates and can thus be ignored. Extending this, suppose we have bounds $\tilde{r}_2(i,f)$ and $\tilde{l}(i,j)$ for $j\in\mathcal{G}$ such that 
$\tilde{r}_2(i,f) > r_2(f)$ and $\tilde{l}(i,j) < \|x(i) - c(j)\|$. Then $\tilde{r}_2(i,f) < \tilde{l}(i,j)$ means that centroid $j$ can be ignored. It remains to define relevant bounds $\tilde{r}_2(i,f)$ and $\tilde{l}(i,j)$.

For $\tilde{r}_2(i,f)$, one keeps track of the second nearest centroid found thus far while looping over the centroids in $\mathcal{G}(f)$. Then for $\tilde{l}(i,j)$ we could take $l(i, f)$, but a better choice is $\tilde{l}(i,j) - q(f) + p(j)$, which replaces the maximum group displacement in the last round with the exact displacement of centroid $j$.

The Yinyang test to determine whether centroid $j$ needs be considered is thus finally, 
\begin{align}
\label{eqn::yybc}
l(i, &f) - q(f) + p(j) > \tilde{r}_2(i,f) \implies \\
&\mbox{ centroid $j$ lies beyond radius $r_2$, can be ignored. } \notag
\end{align}

\subsection{SMN, MSN, MNS}
\label{supp:sec:3pos}
A lower bound to at time $t_0 + \delta_t$ on the distance from $x(i)$ to a group of centroids with group index $f$ can be computed in three different ways. Letting $\Delta_{t_0, \delta_t}$ denote the update term in
\begin{equation*}
l_{t_0 + \delta_t}(i,f) = \min_{j \in \mathcal{G}(i)}\left(\|x(i) - c_{t_0}(j)\|\right) - \Delta_{t_0, \delta_t},
\end{equation*}
the three possibilities are  
\begin{align*}
\Delta_{t_0, \delta_t}^{SMN} &= \sum_{t' = t_0}^{t_0 + \delta t - 1} \max_{j \in \mathcal{G}(i)}{(\|c_{t' + 1}(j) -c_{t'}(j)\| )},\\
\Delta_{t_0, \delta_t}^{MSN} &= \max_{j \in \mathcal{G}(i)} \left(\sum_{t' = t_0}^{t_0 + \delta t - 1}\|c_{t' + 1}(j) -c_{t'}(j)\| \right),\\
\Delta_{t_0, \delta_t}^{MNS} &= \max_{j \in \mathcal{G}(i)} \left(\|c_{t_0 + \delta_t}(j) -c_{t_0}(j)\| \right).
\end{align*}
The term $\Delta_{t_0, \delta_t}^{SMN}$ corresponds to the classic approach used in all previous works. The term $\Delta_{t_0, \delta_t}^{MSN}$ corresponds to an intermendiate where improved bounds can be obtained without storing centroids. The term $\Delta_{t_0, \delta_t}^{MNS}$ corresponds to the approach providing the tightest bounds, and is the one we use throughout.  

\begin{table}[ht!]
\centering
\begin{tabular}{ |c|c|c|c| } 
\hline
& Data set & $d$ & $N$ \\
\hline
i & birch & 2 & 100,000 \\
ii & europe & 2 & 169,300 \\
iii & urand2 & 2 & 1,000,000 \\
iv & ldfpads & 3 & 164,850 \\
v & conflongdemo & 3 & 164,860 \\
vi & skinseg & 4 & 200,000 \\
vii & tsn & 4 & 200,000 \\
viii & colormoments & 9 & 68,040 \\
ix & mv & 11 & 40,760 \\
x & wcomp & 15 & 165,630 \\
xi & house16h & 17 & 22,780 \\
xii & keggnet & 28 & 65,550 \\
xiii & urand30 & 30 & 1,000,000 \\
xiv & mnist50 & 50 & 60,000 \\
xv & miniboone & 50 & 130,060 \\
xvi & covtype & 55 & 581,012 \\
xvii & uscensus & 68 & 2,458,285 \\
xviii & kddcup04 & 74 & 145,750 \\
xix & stl10 & 108 & 1,000,000 \\
xx & gassensor & 128 & 13,910 \\
xxi & kddcup98 & 310 & 95,000 \\
xxii & mnist784 & 784 & 60,000 \\
\hline
\end{tabular}

\caption{Fullnames of the 22 datasets used. All datasets are preprocessed such that features have mean zero and variance 1.}
\label{tab:datasetspec}
\end{table}

\section{Full Results Tables}
\label{sec::supp::results}
\newgeometry{bottom=2cm}

\begin{landscape}
\begin{table}
\centering

\begin{tabular}{ |b|bb|bb|bbbba|bbabaaa|aaba|aabbbaa| }
	
\hline
	
& 
\begin{sideways} mean iterations \, \end{sideways}

& 
\begin{sideways} SD iterations \, \end{sideways}

& 
\begin{sideways} mean fastest [s] \, \end{sideways}

& 
\begin{sideways} SD fastest \, \end{sideways}
		
&		
\begin{sideways}
\texttt{bay-sta} \,
\end{sideways}
		
&		
\begin{sideways}
\texttt{mlp-sta} \,
\end{sideways}
		
&		
\begin{sideways}
\texttt{pow-sta} \,
\end{sideways}
		
&		
\begin{sideways}
\texttt{vlf-sta} \,
\end{sideways}
		
&		
\begin{sideways}
\texttt{own-sta} \,
\end{sideways}
		
&		
\begin{sideways}
\texttt{bay-ham} \,
\end{sideways}
		
&		
\begin{sideways}
\texttt{mlp-ham} \,
\end{sideways}
		
&		
\begin{sideways}
\texttt{own-ham} \,
\end{sideways}
		
&		
\begin{sideways}
\texttt{bay-ann} \,
\end{sideways}
		
&		
\begin{sideways}
\texttt{own-ann} \,
\end{sideways}
		
&		
\begin{sideways}
\texttt{own-exp} \,
\end{sideways}
		
&		
\begin{sideways}
\texttt{own-exp-ns} \,
\end{sideways}
		
&		
\begin{sideways}
\texttt{own-syin} \,
\end{sideways}
		
&		
\begin{sideways}
\texttt{own-syin-ns} \,
\end{sideways}
		
&		
\begin{sideways}
\texttt{pow-yin} \,
\end{sideways}
		
&		
\begin{sideways}
\texttt{own-yin} \,
\end{sideways}
		
&		
\begin{sideways}
\texttt{own-selk} \,
\end{sideways}
		
&		
\begin{sideways}
\texttt{own-selk-ns} \,
\end{sideways}
		
&		
\begin{sideways}
\texttt{bay-elk} \,
\end{sideways}
		
&		
\begin{sideways}
\texttt{mlp-elk} \,
\end{sideways}
		
&		
\begin{sideways}
\texttt{vlf-elk} \,
\end{sideways}
		
&		
\begin{sideways}
\texttt{own-elk} \,
\end{sideways}
		
&		
\begin{sideways}
\texttt{own-elk-ns} \,
\end{sideways}

\\\hline
	i  & 139 & 35.4 & 0.26 & 0.04    & 62.2  & 44.7  & 105  & 33.7  & 19.2  & 5.75  & 4.87  & 4.67  & \underline{1.72}  & 2.19  & 1.05  & \bf{1.00}  & 1.91  & 2.01  & 14.9  & 1.98  & 9.37  & 8.42  & 14.8  & 11.1  & 21.3  & 7.53  & 5.55  \\ 
ii  & 335 & 132 & 1.69 & 0.25    & 38.2  & 28.1  & 60.3  & 21.5  & 10.3  & 6.44  & 5.00  & 5.05  & \underline{1.71}  & 1.99  & 1.07  & \bf{1.00}  & 3.05  & 3.13  & 14.6  & 2.96  & 6.08  & 5.60  & 10.8  & 8.70  & 15.9  & 6.94  & 6.70  \\ 
iii  & 455 & 187 & 5.23 & 1.47    & 100  & 76.4  & 153  & 56.4  & 29.9  & 4.00  & 4.25  & 3.35  & \underline{1.75}  & 1.99  & 1.05  & \bf{1.00}  & 2.04  & 1.95  & 23.0  & 2.33  & 15.1  & 13.6  & 26.6  & 19.1  & 34.1  & 12.3  & 8.67  \\ 
iv  & 270 & 96.3 & 2.17 & 0.48    & 29.0  & 17.9  & 37.7  & 15.2  & 6.79  & 3.99  & 2.90  & 2.51  & \underline{2.03}  & 1.64  & 1.04  & \bf{1.00}  & 1.02  & 1.06  & 6.78  & 1.09  & 3.66  & 3.30  & 7.09  & 6.68  & 11.0  & 4.54  & 3.80  \\ 
v  & 277 & 73.6 & 2.35 & 0.48    & 27.5  & 18.1  & 35.7  & 14.4  & 6.80  & 4.07  & 3.06  & 2.55  & \underline{2.05}  & 1.66  & 1.04  & \bf{1.00}  & 1.01  & 1.05  & 6.48  & 1.08  & 3.47  & 3.14  & 6.72  & 6.57  & 10.5  & 4.31  & 3.63  \\ 
vi  & 235 & 107 & 1.76 & 0.61    & 44.3  & 25.7  & 55.6  & 17.7  & 9.96  & 4.68  & 3.34  & 2.63  & \underline{2.23}  & 1.69  & 1.05  & \bf{1.00}  & 1.20  & 1.21  & 8.45  & 1.32  & 4.76  & 4.29  & 8.79  & 8.22  & 13.4  & 5.16  & 4.20  \\ 
vii  & 117 & 25.2 & 1.32 & 0.18    & 28.9  & 15.9  & 34.2  & 11.7  & 5.85  & 4.29  & 2.83  & 2.42  & \underline{2.61}  & 1.76  & 1.25  & 1.19  & 1.02  & \bf{1.00}  & 5.95  & 1.06  & 3.24  & 2.89  & 5.72  & 4.90  & 8.82  & 3.19  & 2.60  \\ 
viii  & 203 & 48.4 & 1.19 & 0.20    & 31.4  & 14.1  & 26.3  & 10.3  & 5.61  & 6.47  & \underline{3.33}  & 2.42  & 5.08  & 2.14  & 2.39  & 2.24  & 1.02  & \bf{1.00}  & 5.11  & 1.28  & 2.25  & 2.02  & 6.22  & 6.38  & 10.0  & 4.10  & 3.09  \\ 
ix  & 127 & 35.4 & 0.49 & 0.07    & 32.6  & 14.7  & 25.5  & 12.0  & 5.02  & 7.80  & \underline{4.02}  & 2.66  & 7.99  & 2.75  & 2.63  & 2.51  & 1.02  & \bf{1.00}  & 4.83  & 1.32  & 2.12  & 1.89  & 6.12  & 6.12  & 9.77  & 3.43  & 2.37  \\ 
x  & 130 & 27.2 & 2.16 & 0.46    & 38.7  & 17.2  & 29.2  & 12.0  & 5.11  & 10.7  & 5.21  & 2.97  & 4.90  & 1.83  & 1.23  & 1.21  & \bf{1.00}  & 1.03  & 4.56  & 1.39  & 1.89  & 1.67  & 3.29  & \underline{3.28}  & 5.46  & 2.00  & 1.67  \\ 
xi  & 111 & 16.6 & 0.52 & 0.07    & 20.8  & 9.62  & 14.0  & 5.36  & 2.82  & 9.34  & 4.89  & 2.40  & 5.96  & 1.88  & 2.33  & 2.22  & 1.11  & 1.07  & 3.75  & 1.73  & 1.08  & \bf{1.00}  & \underline{2.90}  & 3.30  & 4.34  & 2.07  & 1.90  \\ 
xii  & 113 & 33.4 & 0.92 & 0.14    & 53.3  & 23.3  & 30.2  & 10.3  & 4.86  & 12.0  & 4.21  & 2.11  & 5.58  & 1.47  & 1.82  & 1.75  & \bf{1.00}  & 1.00  & 4.36  & 1.74  & 1.66  & 1.46  & 5.26  & \underline{3.64}  & 4.84  & 1.95  & 1.60  \\ 
xiii  & 2576 & 110 & 156 & 1.86    & t  & t  & t  & t  & t  & 9.15  & \underline{4.61}  & 1.85  & 9.26  & 1.97  & 1.82  & 1.70  & 1.04  & \bf{1.00}  & 6.06  & 1.58  & 3.37  & 3.04  & 4.70  & 5.40  & t  & 3.46  & 3.16  \\ 
xiv  & 152 & 58.3 & 1.72 & 0.33    & 67.4  & 21.6  & 29.3  & 10.7  & 5.60  & 17.8  & 5.76  & 2.47  & 18.2  & 2.61  & 2.44  & 2.34  & 1.01  & \bf{1.00}  & 4.00  & 2.02  & 1.47  & 1.25  & 3.04  & \underline{2.66}  & 4.72  & 1.74  & 1.39  \\ 
xv  & 376 & 90.1 & 7.18 & 1.13    & 87.5  & 27.2  & 37.9  & 13.7  & 6.97  & 25.4  & 7.84  & 3.40  & 15.3  & 2.44  & 3.27  & 3.09  & \bf{1.00}  & 1.06  & 4.53  & 2.05  & 1.54  & 1.38  & \underline{4.20}  & 4.22  & 5.72  & 2.84  & 2.76  \\ 
xvi  & 127 & 35.0 & 8.45 & 0.94    & 120  & 38.9  & 51.4  & 20.3  & 10.8  & 18.1  & 6.34  & 2.67  & 15.2  & 2.51  & 1.92  & 1.85  & \bf{1.00}  & 1.04  & 4.72  & 2.05  & 1.85  & 1.60  & \underline{3.38}  & 4.01  & 5.73  & 2.01  & 1.58  \\ 
xvii  & 102 & 38.6 & 43.9 & 6.73    & t  & 28.3  & m  & 14.0  & 6.09  & 19.5  & 6.11  & 2.55  & 17.5  & 2.47  & 2.34  & 2.26  & 1.00  & \bf{1.00}  & m  & 2.29  & 1.37  & m  & \underline{3.27}  & 3.44  & 4.77  & 1.83  & m  \\ 
xviii  & 334 & 118 & 13.5 & 3.46    & 63.9  & 18.7  & 26.0  & 9.58  & 4.85  & 23.8  & 6.90  & 2.89  & 22.3  & 2.90  & 2.86  & 2.81  & 1.14  & 1.12  & 4.63  & 2.75  & 1.13  & \bf{1.00}  & 2.51  & \underline{1.91}  & 3.51  & 1.31  & 1.16  \\ 
xix  & 408 & 145 & 92.5 & 27.0    & t  & t  & t  & t  & 7.18  & t  & 5.77  & 2.29  & t  & 2.38  & 2.16  & 2.10  & 1.05  & 1.10  & 4.35  & 2.92  & 1.14  & \bf{1.00}  & \underline{2.59}  & 2.74  & 3.85  & 1.58  & 1.60  \\ 
xx  & 57.3 & 15.9 & 0.23 & 0.05    & 101  & 28.2  & 40.0  & 14.6  & 6.73  & 33.7  & 9.58  & 3.65  & 15.8  & 2.27  & 3.37  & 3.25  & 1.72  & 1.78  & 6.14  & 4.56  & 1.12  & 1.13  & 2.53  & \underline{2.21}  & 2.41  & \bf{1.00}  & 1.02  \\ 
xxi  & 178 & 58.8 & 9.96 & 1.97    & 121  & 30.1  & 44.0  & 16.7  & 7.98  & 41.4  & 10.5  & 3.72  & 42.1  & 3.94  & 3.66  & 3.57  & 1.47  & 1.47  & 5.76  & 4.64  & 1.18  & \bf{1.00}  & 3.10  & \underline{2.07}  & 2.39  & 1.32  & 1.14  \\ 
xxii  & 131 & 25.9 & 9.89 & 1.23    & 143  & 34.8  & 50.1  & 20.0  & 10.0  & 40.1  & 10.2  & 3.97  & 40.6  & 4.10  & 3.87  & 3.75  & 1.29  & 1.34  & 4.23  & 3.64  & 1.25  & \bf{1.00}  & 3.63  & 1.82  & \underline{1.71}  & 1.27  & 1.03  \\ 

\hline
\end{tabular}

\caption{Results with $k$ = 100 by dataset (rows). Columns 6 to the end contain mean times over the \nseeds{} initialisations, relative to the fastest algorithm, that is the algorithm with the lowest mean time, corresponding to the entry $\mathbf{1.00}$. The mean and standard deviation of the number of iterations to convergence are given in columns 2 and 3. The mean and standard deviation of the time of the fastest algorithm are given in columns 4 and 5. `t' and `m' correspond to timeout (\timelimit{}) and memory (\memlimit{}) failures respectively. The fastest implementation for all data sets is always \texttt{own}. The fastest non-\texttt{own} implementation for each data set is underlined, where non-\texttt{own} implementations correspond to white columns.}
\label{supp:tab:100}
\end{table}
\end{landscape}

\begin{landscape}
\begin{table}
\centering

\begin{tabular}{ |b|bb|bb|bbbba|bbabaaa|aaba|aabbbaa| }
	
\hline
	
& 
\begin{sideways} mean iterations \, \end{sideways}

& 
\begin{sideways} SD iterations \, \end{sideways}

& 
\begin{sideways} mean fastest [s] \, \end{sideways}

& 
\begin{sideways} SD fastest \, \end{sideways}
		
&		
\begin{sideways}
\texttt{bay-sta} \,
\end{sideways}
		
&		
\begin{sideways}
\texttt{mlp-sta} \,
\end{sideways}
		
&		
\begin{sideways}
\texttt{pow-sta} \,
\end{sideways}
		
&		
\begin{sideways}
\texttt{vlf-sta} \,
\end{sideways}
		
&		
\begin{sideways}
\texttt{own-sta} \,
\end{sideways}
		
&		
\begin{sideways}
\texttt{bay-ham} \,
\end{sideways}
		
&		
\begin{sideways}
\texttt{mlp-ham} \,
\end{sideways}
		
&		
\begin{sideways}
\texttt{own-ham} \,
\end{sideways}
		
&		
\begin{sideways}
\texttt{bay-ann} \,
\end{sideways}
		
&		
\begin{sideways}
\texttt{own-ann} \,
\end{sideways}
		
&		
\begin{sideways}
\texttt{own-exp} \,
\end{sideways}
		
&		
\begin{sideways}
\texttt{own-exp-ns} \,
\end{sideways}
		
&		
\begin{sideways}
\texttt{own-syin} \,
\end{sideways}
		
&		
\begin{sideways}
\texttt{own-syin-ns} \,
\end{sideways}
		
&		
\begin{sideways}
\texttt{pow-yin} \,
\end{sideways}
		
&		
\begin{sideways}
\texttt{own-yin} \,
\end{sideways}
		
&		
\begin{sideways}
\texttt{own-selk} \,
\end{sideways}
		
&		
\begin{sideways}
\texttt{own-selk-ns} \,
\end{sideways}
		
&		
\begin{sideways}
\texttt{bay-elk} \,
\end{sideways}
		
&		
\begin{sideways}
\texttt{mlp-elk} \,
\end{sideways}
		
&		
\begin{sideways}
\texttt{vlf-elk} \,
\end{sideways}
		
&		
\begin{sideways}
\texttt{own-elk} \,
\end{sideways}
		
&		
\begin{sideways}
\texttt{own-elk-ns} \,
\end{sideways}

\\\hline
	i  & 120 & 20.5 & 2.80 & 0.32    & 48.0  & 32.0  & 104  & 25.4  & 12.0  & 12.5  & 8.50  & 8.29  & \underline{1.49}  & 1.41  & 1.01  & \bf{1.00}  & 1.01  & 1.27  & 8.53  & 1.13  & 6.83  & 6.32  & 11.5  & 17.5  & 16.3  & 5.54  & 4.32  \\ 
ii  & 533 & 66.7 & 12.1 & 1.22    & 82.6  & 56.0  & t  & 43.7  & 21.6  & 19.0  & 13.0  & 12.5  & \underline{2.05}  & 1.76  & 1.01  & \bf{1.00}  & 1.62  & 2.10  & 15.2  & 1.90  & 11.6  & 11.1  & 19.9  & 36.2  & 26.6  & 13.0  & 11.9  \\ 
iii  & 406 & 86.5 & 19.9 & 2.10    & t  & t  & t  & t  & 56.5  & 15.6  & 11.1  & 10.6  & \underline{2.22}  & 2.13  & 1.02  & \bf{1.00}  & 3.05  & 2.48  & 28.5  & 3.33  & m  & m  & m  & m  & m  & m  & m  \\ 
iv  & 193 & 46.5 & 7.22 & 1.07    & 60.4  & 34.2  & 95.4  & 30.2  & 13.5  & 17.8  & 11.0  & 10.0  & \underline{4.49}  & 2.83  & 1.01  & \bf{1.00}  & 1.15  & 1.34  & 9.19  & 1.32  & 6.93  & 6.43  & 12.0  & 24.9  & 18.6  & 7.51  & 6.32  \\ 
v  & 197 & 25.2 & 7.16 & 0.55    & 62.1  & 35.1  & 98.1  & 31.1  & 13.7  & 17.3  & 10.7  & 9.79  & \underline{4.41}  & 2.77  & 1.02  & \bf{1.00}  & 1.15  & 1.34  & 9.17  & 1.31  & 7.10  & 6.61  & 12.4  & 25.3  & 19.3  & 7.65  & 6.39  \\ 
vi  & 287 & 87.0 & 10.6 & 2.04    & 87.0  & 43.8  & 132  & 33.4  & 15.1  & 21.7  & 12.3  & 10.5  & \underline{4.39}  & 2.46  & 1.03  & \bf{1.00}  & 1.24  & 1.39  & 10.4  & 1.43  & 8.42  & 7.78  & 14.7  & 30.7  & 23.2  & 9.13  & 7.45  \\ 
vii  & 94.0 & 21.8 & 4.17 & 0.41    & 71.7  & 36.2  & 105  & 28.0  & 12.6  & 15.6  & 9.18  & 7.83  & \underline{4.85}  & 2.80  & 1.02  & \bf{1.00}  & 1.25  & 1.37  & 9.10  & 1.39  & 7.23  & 6.71  & 11.9  & 19.2  & 18.3  & 6.08  & 4.86  \\ 
viii  & 87.9 & 17.1 & 3.10 & 0.33    & 50.8  & 20.9  & 50.3  & 16.3  & 6.43  & 25.0  & 11.5  & 8.33  & 15.1  & 5.10  & 5.18  & 4.90  & 1.16  & \bf{1.00}  & \underline{6.69}  & 1.45  & 3.24  & 3.09  & 7.17  & 16.5  & 11.8  & 5.40  & 4.23  \\ 
ix  & 51.1 & 5.74 & 1.28 & 0.07    & 48.7  & 20.8  & 44.6  & 17.6  & 6.10  & 24.3  & 11.7  & 7.40  & 23.0  & 6.90  & 5.01  & 4.80  & 1.21  & \bf{1.00}  & \underline{6.13}  & 1.51  & 2.92  & 2.73  & 6.39  & 15.2  & 9.73  & 4.30  & 3.51  \\ 
x  & 201 & 41.6 & 12.5 & 1.30    & 102  & 42.0  & 90.3  & 30.7  & 11.3  & 50.6  & 22.7  & 12.9  & 12.6  & 3.54  & 1.96  & 1.91  & \bf{1.00}  & 1.11  & \underline{6.93}  & 1.36  & 4.28  & 3.95  & 7.80  & 16.9  & 13.1  & 5.41  & 4.64  \\ 
xi  & 46.6 & 8.49 & 1.17 & 0.07    & 37.7  & 15.7  & 29.1  & 9.82  & 3.79  & 29.2  & 14.2  & 7.16  & 17.8  & 4.42  & 5.75  & 5.61  & 1.23  & \bf{1.00}  & \underline{4.97}  & 1.72  & 1.73  & 1.66  & 5.48  & 9.12  & 6.47  & 3.32  & 2.77  \\ 
xii  & 32.8 & 3.81 & 1.87 & 0.07    & 73.4  & 31.5  & 47.5  & 15.0  & 6.86  & 28.9  & 8.55  & 4.36  & 9.20  & 2.09  & 1.99  & 1.95  & 1.04  & \bf{1.00}  & \underline{4.60}  & 1.37  & 2.17  & 2.03  & 12.8  & 6.60  & 5.04  & 2.07  & 1.57  \\ 
xiii  & 738 & 108 & 382 & 31.1    & t  & t  & t  & t  & t  & t  & t  & t  & t  & t  & t  & t  & 1.14  & \bf{1.00}  & t  & 1.87  & m  & m  & m  & m  & m  & m  & m  \\ 
xiv  & 58.9 & 7.76 & 5.05 & 0.21    & 88.2  & 26.8  & 43.5  & 14.0  & 5.87  & 55.6  & 16.9  & 8.14  & 57.4  & 8.18  & 7.97  & 7.67  & 1.36  & \bf{1.00}  & \underline{5.15}  & 2.47  & 1.75  & 1.46  & 5.77  & 5.68  & 6.33  & 3.05  & 1.96  \\ 
xv  & 181 & 41.5 & 15.8 & 1.93    & t  & 58.0  & 92.7  & 29.6  & 14.4  & 109  & 33.1  & 15.5  & 59.1  & 8.08  & 11.0  & 10.6  & 1.07  & \bf{1.00}  & \underline{5.77}  & 1.84  & 2.60  & 2.49  & 7.56  & 14.6  & 9.56  & 5.30  & 4.14  \\ 
xvi  & 224 & 55.8 & 46.6 & 4.86    & t  & t  & t  & t  & t  & t  & 29.0  & 13.9  & t  & 9.61  & 2.98  & 2.88  & 1.03  & \bf{1.00}  & \underline{6.18}  & 1.55  & m  & m  & m  & m  & m  & m  & m  \\ 
xvii  & 145 & 32.9 & 249 & 11.2    & t  & t  & m  & t  & t  & t  & t  & t  & t  & t  & t  & 7.87  & \bf{1.00}  & m  & m  & 1.73  & m  & m  & m  & m  & m  & m  & m  \\ 
xviii  & 114 & 11.9 & 33.7 & 1.18    & t  & 24.7  & 38.8  & 13.1  & 6.47  & 59.0  & 16.6  & 8.16  & 58.6  & 7.89  & 8.20  & 8.00  & 1.33  & \bf{1.00}  & 5.11  & 2.80  & 1.47  & 1.27  & 3.99  & \underline{3.13}  & 4.30  & 2.24  & 1.71  \\ 
xix  & 612 & 160 & 587 & 76.3    & t  & t  & t  & t  & t  & t  & t  & t  & t  & t  & t  & t  & 1.09  & \bf{1.00}  & t  & 2.63  & m  & m  & m  & m  & m  & m  & m  \\ 
xx  & 18.0 & 1.00 & 0.71 & 0.03    & 98.1  & 27.1  & 42.4  & 15.6  & 10.2  & 61.6  & 16.7  & 7.37  & 20.8  & 2.65  & 2.13  & 2.17  & 1.29  & 1.24  & 4.47  & 2.17  & \bf{1.00}  & 1.06  & 8.31  & 3.50  & \underline{2.32}  & 1.35  & 1.38  \\ 
xxi  & 76.1 & 14.4 & 31.7 & 2.73    & t  & 39.9  & t  & 23.1  & 7.92  & t  & 26.0  & 10.0  & t  & 10.1  & 9.98  & 9.63  & 1.75  & 1.38  & 6.41  & 4.85  & 1.39  & \bf{1.00}  & 5.51  & 3.44  & \underline{2.90}  & 1.92  & 1.37  \\ 
xxii  & 54.8 & 11.0 & 23.0 & 1.90    & t  & 64.5  & t  & 38.1  & 13.8  & t  & 42.4  & 15.3  & t  & 14.8  & 14.6  & 14.2  & 2.10  & 1.52  & 6.49  & 5.53  & 1.82  & \bf{1.00}  & 9.85  & 3.56  & \underline{2.61}  & 2.04  & 1.22  \\ 

\hline
\end{tabular}

\caption{As per Table~\ref{supp:tab:100}, but with $k$ = 1000}
\label{supp:tab:1000}
\end{table}
\end{landscape}

\end{document}